\documentclass[runningheads]{llncs}

\usepackage{eccv}

\usepackage{eccvabbrv}

\usepackage{epsfig}
\usepackage{graphicx}
\usepackage{amsmath}
\usepackage{amssymb}
\usepackage{booktabs}
\usepackage{bm}
\usepackage{amsfonts}
\usepackage{mathrsfs}
\usepackage{pifont}
\usepackage{float}
\usepackage{subcaption}
\usepackage{indentfirst}
\usepackage{algorithm}
\usepackage{algorithmic}
\usepackage{bbding}
\usepackage{multicol}

\usepackage[accsupp]{axessibility}  

\usepackage{hyperref}
\usepackage[capitalize]{cleveref}
\crefname{section}{Sec.}{Secs.}
\Crefname{section}{Section}{Sections}
\Crefname{table}{Table}{Tables}
\crefname{table}{Tab.}{Tabs.}

\usepackage{orcidlink}

\begin{document}

\title{KFD-NeRF: Rethinking Dynamic NeRF with Kalman Filter}


\author{
Yifan Zhan\inst{1}
\and
Zhuoxiao Li\inst{1}
\and
Muyao Niu\inst{1}
\and
Zhihang Zhong\inst{3}
\and
Shohei Nobuhara\inst{2}
\and
Ko Nishino\inst{2}
\and
Yinqiang Zheng\thanks{\footnotesize denotes corresponding author.}\inst{1}
}

\authorrunning{Y.Zhan, Z.Li et al.}

\institute{
The University of Tokyo
\and
Kyoto University
\and
Shanghai Artificial Intelligence Laboratory
}

\maketitle

\begin{abstract}
  We introduce KFD-NeRF, a novel dynamic neural radiance field integrated with an efficient and high-quality motion reconstruction framework based on Kalman filtering. Our key idea is to model the dynamic radiance field as a dynamic system whose temporally varying states are estimated based on two sources of knowledge: observations and predictions. We introduce a novel plug-in Kalman filter guided deformation field that enables accurate deformation estimation from scene observations and predictions. We use a shallow Multi-Layer Perceptron (MLP) for observations and model the motion as locally linear to calculate predictions with motion equations. To further enhance the performance of the observation MLP, we introduce regularization in the canonical space to facilitate the network's ability to learn warping for different frames. Additionally, we employ an efficient tri-plane representation for encoding the canonical space, which has been experimentally demonstrated to converge quickly with high quality. This enables us to use a shallower observation MLP, consisting of just two layers in our implementation. We conduct experiments on synthetic and real data and compare with past dynamic NeRF methods. Our KFD-NeRF demonstrates similar or even superior rendering performance within comparable computational time and achieves state-of-the-art view synthesis performance with thorough training. Github page: \url{https://github.com/Yifever20002/KFD-NeRF}.
\end{abstract}

\section{Introduction}
\label{sec:Introduction}

Neural Radiance Fields (NeRF)~\cite{mildenhall2021nerf} have demonstrated outstanding success as a versatile and accurate 3D representation of real-world scenes, which has led to its wide adoption in daily and industrial applications in numerous domains. One of the remaining key desiderata for NeRFs is its extension to dynamic scenes. 

Existing dynamic NeRF methods can be broadly categorized into two approaches. One is to learn deformation fields for motion warping (\eg, D-NeRF~\cite{pumarola2021d} and TiNeuVox~\cite{fang2022fast}). Another is to disregard motion priors and directly interpolate time in the feature space (\eg, DyNeRF~\cite{li2022neural} and KPlanes~\cite{fridovich2023k}). These approaches, however, often overlook the characteristics of a dynamic radiance field as a time-state sequence, missing the opportunity to fully leverage temporal contextual information. 

\begin{figure}[tb]
  \centering
  \includegraphics[width=1\linewidth]{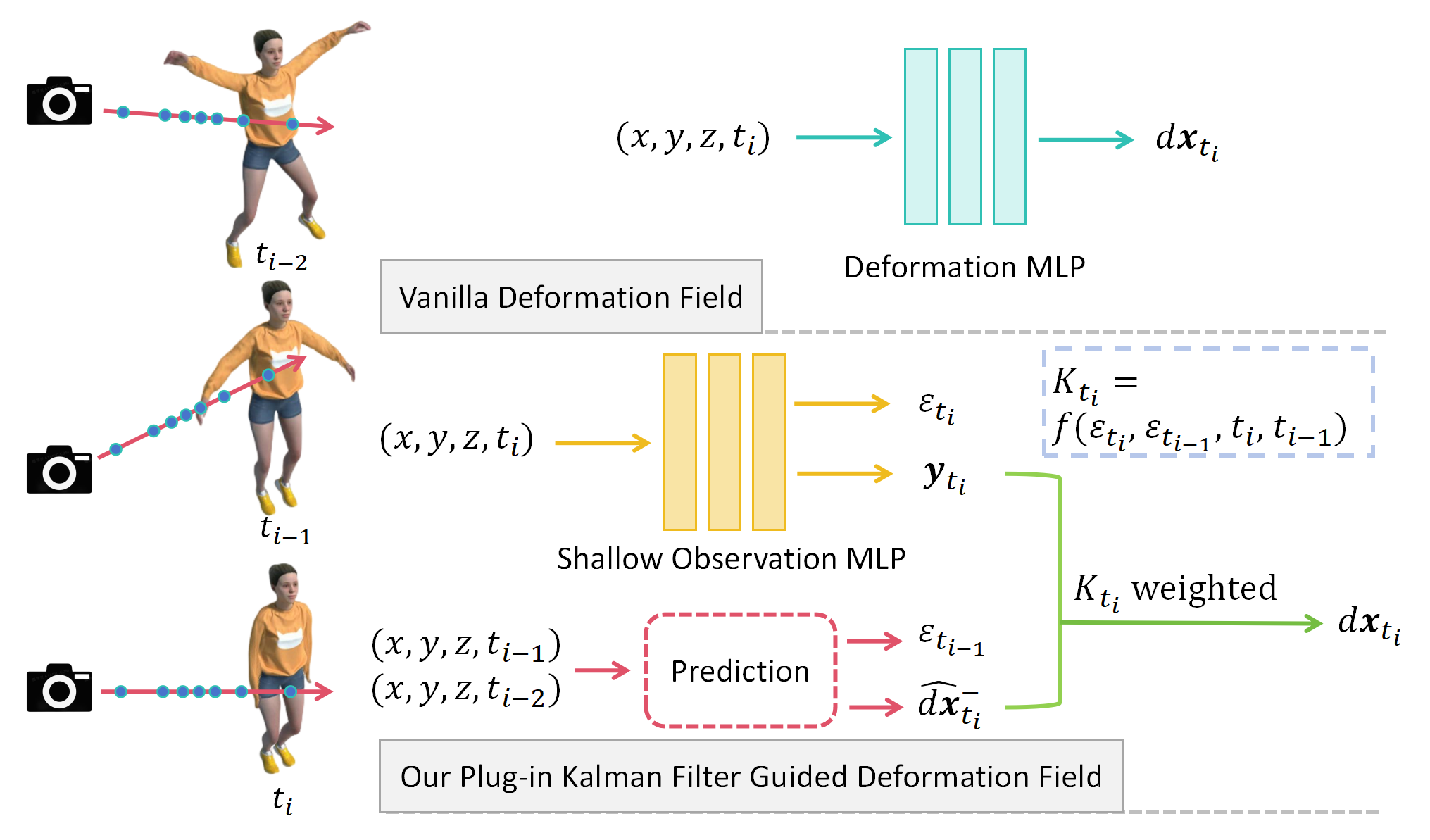}
      \caption{In contrast to a vanilla deformation field, our plug-in Kalman filter guided deformation field consists of a prediction branch along with the direct observations from input data. From noise related terms $\varepsilon_{t_i}$ and $\varepsilon_{t_{i-1}}$ we learn Kalman gain $K_{t_i}$, weighting observations $y_{t_{i}}$ and predictions $\hat{d\bm{x}}_{t_i}^-$ for more accurate deformation estimations.}
  \label{fig:introduction_01}
\end{figure}

In this work, we draw inspirations from control theory and model the 4D radiance field as a dynamic system with temporally-varying states. Our state estimates of this dynamic system come from two sources of knowledge: observations based on the input data and predictions based on the system physical dynamics. Optimal state estimates cannot be achieved with only one of these two sources of knowledge. On the one hand, observations, as commonly used in previous dynamic NeRF works, are inherently error-prone due to the discrete temporal sampling of the dynamic scenes. On the other hand, predictions are governed by the correctness of the assumed kinematic model and may struggle to maintain accuracy for real dynamic scenes.

To maximize combined potential of both observations and predictions, we introduce an efficient plug-and-play Kalman filter~\cite{welch1995introduction} module to optimize state estimations of our dynamic system. In \cref{fig:introduction_01} we illustrate our plug-in Kalman filter guided deformation field. We model the 4D radiance field as a single-state system, with the state denoted as $d\bm{x}_{t_i}$ for the current frame deformation. In contrast to a vanilla deformation field that only considers system observations in the current frame, our approach incorporates richer information from previous frames by introducing a prediction branch based on a motion equation. Given the absence of prior trajectory regarding the scene's motion, we employ local linear motion.

Both observations and predictions are weighted using a learnable Kalman gain to calculate precise deformation estimations. During the initial stages of training, predictions primarily influence the process, facilitating the convergence of frames with substantial motion. In the later stages of training, observations take precedence, allowing for the recovery of more precise and fine-grained motion details.

All points in the real space are warped to a time-independent canonical space according to the estimated deformations. To further improve the performance of our two-branch deformable model, we employ an efficient tri-plane spatial representation for encoding our canonical space. Experimental evidence shows that this representation permits a shallower observation MLP with only two layers in our implementation. At the same time, we improve the warping ability of the observation MLP by regularizing the learning of the radiance field in the canonical space. 

In summary, our contributions are as follows:

\begin{itemize}

  \item [1)]
    The first method for modeling 4D radiance fields as dynamic systems by integrating a Kalman filter into the deformation field formulation, which results in a plug-in, efficient method for estimating deformations;
  \item [2)]
    KFD-NeRF, a novel deformable NeRF with the Kalman filter plug-in and a tri-plane spatial representation, trained with a novel strategy of gradually releasing temporal information to facilitate learning dynamic systems;
 \item [3)]
    and regularization in the canonical space for enhancing the learning capacity of a shallow observation MLP. We achieve state-of-the-art results on both synthetic and real data with all these designs, compared to dynamic NeRFs.
\end{itemize}

\section{Related Works}
\label{sec:Related Works}

\paragraph{Neural Radiance Fields Representation.} NeRF~\cite{mildenhall2021nerf} represents 3D scenes based on an implicit representation encoded by an MLP. Given multi-view images and corresponding camera poses as input, the MLP is trained by ``analysis by synthesis,'' achieving novel view synthesis and coarse geometry reconstruction. Vanilla NeRF leaves room for improvement that has attracted abundant research. 

Many methods~\cite{liu2020neural,yu2021plenoctrees,sun2022direct,hu2022efficientnerf,fridovich2022plenoxels} use sparse voxel grids to represent the 3D scenes. While grid-based modeling achieves fast training and rendering, fine details require high-resolution grids leading to excessive memory consumption. The optimization process for grids is also unstable, failing to leverage the smoothness bias brought by MLP. Instant-NGP~\cite{muller2022instant} proposes an efficient hash encoding for mapping spatial and feature domains. For sparsely-observed regions, however, this one-to-many mapping can easily introduce noise. The tri-plane representation~\cite{peng2020convolutional,chan2022efficient,chen2022tensorf} significantly reduces the memory footprint by leveraging its low-rank decomposition. We use this representation to build a fine-detailed canonical space.

\paragraph{Dynamic Neural Rendering.} For dynamic scenarios, modeling the time dimension together with the spatial representation becomes the main challenge. One approach to dynamic NeRF~\cite{gao2021dynamic,li2022neural,li2021neural,xian2021space,fridovich2023k,cao2023hexplane,liu2023robust,park2023temporal,jang2022d,gan2023v4d,shao2023tensor4d} is to add timestamps as an extra dimension to the 3D space and formulate 4D interpolation. Though intuitive, this leads to temporal incoherence due to the lack of supervision and priors between frames. Another approach~\cite{pumarola2021d,park2021nerfies,fang2022fast,du2021neural,li2021neural,tretschk2021non,yuan2021star,park2021hypernerf,choe2023spacetime,guo2023forward, guo2022neural} is to model the motion with deformation fields that warp points in arbitrary frames to the canonical space conditioned on the timestamps. This design can take advantage of the MLP smoothness bias in accordance with the motion smoothness prior. Nevertheless, deformation fields fail to find correspondences in frames with significant motion. We emphasize the performance improvement resulting from the switch in backbone spatial representation from D-NeRF~\cite{pumarola2021d} (MLP-based) to TiNeuVox~\cite{fang2022fast} (voxel grids based) and later analyze the impact of different spatial representations on the deformation field. We also compare with latest point-based 4D rendering works~\cite{wu20234d,yang2023deformable,luiten2023dynamic} inspired by 3D Gaussian Splatting~\cite{kerbl20233d}.

\paragraph{Deep Kalman Filter.} The combination of deep learning and Kalman filtering has been explored to address the challenges of incomplete observations in various scenarios. Some works~\cite{lu2018deep, zhou2020kfnet} use Convolutional Neural Network (CNN)-based Kalman filtering for video compression and camera localization. Recurrent Neural Network (RNN)-based Kalman filter has also been used in some studies~\cite{revach2022kalmannet, coskun2017long} to improve state estimation and to optimize pose regularization. The Transformer in~\cite{zhang2023transformer} enables a more comprehensive exploitation of temporal contextual information. All of these works along with~\cite{haarnoja2016backprop, fraccaro2017disentangled, becker2019recurrent}, however, model dynamics only in the learned latent space without taking into account physical priors, with the exception of \cite{guen2020disentangling} which fuses known physical priors and network outputs to construct a Kalman filter for video prediction.
In this paper, for the first time, we employ a neural Kalman filter to assist in the dynamic NeRF task. We derive a two-stream method consisting of a shallow MLP and physical priors to achieve efficient motion estimation.

\section{Preliminaries}
\label{sec:Preliminaries}

\subsection{NeRF and Volume Rendering Revisited}

NeRF~\cite{mildenhall2021nerf} consists of three parts: sampling, volume mapping, and rendering. In the sampling process, points $\bm{x}\in\mathbb{R}^3$ are sampled along rays calculated from the camera pose. Then, in the volume mapping process, each $\bm{x}$ in the 3D world, as well as its viewing direction $\bm{v}\in\mathbb{R}^3$, is queried to output the volume density $\sigma$ and radiance $\bm{c}=(r,g,b)$ of $\bm{x}$. Finally, in the rendering process, the color of each ray is computed with volume rendering~\cite{kajiya1984ray}. The expected color of the ray $\bm{r}(s)=\bm{o}+s\bm{v}$ becomes~\cite{mildenhall2021nerf} 
\begin{equation}
  C(\bm{r})=\int_{s_n}^{s_f}\ T(s)\sigma(\bm{r}(s))\bm{c}(\bm{r}(s),\bm{v})ds\,,
  \label{eq:Rendering equation}
\end{equation}
where $s_n$ and $s_f$ are the near and far bounds, respectively, and 
\begin{equation}
    T(s)=\exp\left(-\int_{s_n}^{s}\ \sigma(\bm{r}(k))dk\right)\,.
\end{equation}

The only supervision for training NeRF comes from the ground truth color $C_{gt}(\bm{r})$ of each ray
\begin{equation}
  \mathcal{L}_{image}=\sum_{\bm{r}\in\mathcal{R}}\ \left\|C(\bm{r})-C_{gt}(\bm{r})\right\|_2^2\,,
  \label{eq:loss function}
\end{equation}
where $\mathcal{R}$ is the ray batch.

\subsection{Kalman Filter}

Consider a dynamic system with input $u_t$, output $y_t$, process noise $n_t$ and measurement noise $m_t$. We want to obtain its state $x_t$ at each frame $t$, assuming the state equation follows
\begin{equation}
  x_t=Ax_{t-1}+Bu_t+n_t\,,
  \label{eq:state equation}
\end{equation}
and the output equation follows
\begin{equation}
  y_t=Cx_t+m_t\,,
  \label{eq:output equation}
\end{equation}
where $A$, $B$ and $C$ are the system matrix, control matrix and output matrix, respectively. 

There are two methods for obtaining the system's state $x_t$ at any given $t$: observation and prediction. The observation method tries to rely on $y_t$ in \cref{eq:output equation} while the prediction method tries to mathematically model the system dynamics to predict the state. Both of these methods, however, come with non-negligible errors due to the existence of $m_t$ and $n_t$ and for inaccurate system modeling.

The Kalman filter algorithm posits that the state $x_t$ is obtained by a weighted sum of the prediction based on the state $x_{t-1}$ and the observation $y_t$. For a specific $t$, estimating the state $x_t$ can be divided into two steps: prediction step and update step. Assuming that the process noise and measurement noise follow zero-mean Gaussian distributions with variances $Q$ and $R$, respectively, at the prediction step, it first calculates a predicted state based on the estimated state at $t-1$
\begin{equation}
  \hat{x}_t^{-}=A\hat{x}_{t-1}+Bu_t\,,
  \label{eq:predicted state estimation}
\end{equation}
and its error covariance
\begin{equation}
  P_t^{-}=AP_{t-1}A^T+Q\,.
  \label{eq:predicted state estimation error covariance}
\end{equation}
In the update step, this predicted state is combined with observation $y_t$ to obtain updated state estimate
\begin{equation}
  \hat{x}_t=\hat{x}_t^{-}+K_t(y_t-C\hat{x}_t^{-})\,,
  \label{eq:updated state estimation}
\end{equation}
and its error covariance
\begin{equation}
  P_t=(I-K_tC)P_t^{-}\,,
  \label{eq:updated state estimation error covariance}
\end{equation}
where 
\begin{equation}
  K_t=\frac{P_t^{-}C^T}{CP_t^{-}C^T+R}
  \label{eq:Kaiman gain}
\end{equation}
is the Kalman gain, reflecting the weights assigned to the observation and prediction components.

\begin{figure}[tb]
  \centering
  \includegraphics[width=1\linewidth]{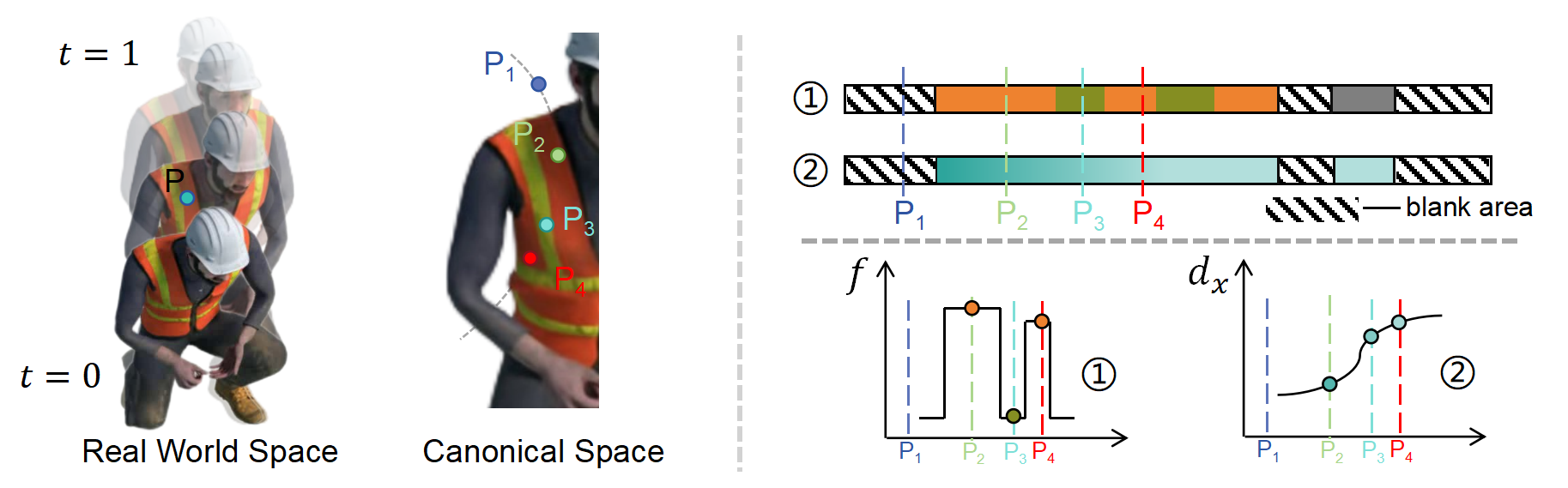}
      \caption{Visualization of feature planes learned by the feature interpolation method \ding{192} and the deformation fields method \ding{193}. We show a point $P$ in the real world space and its corresponding four points $P_1$, $P_2$, $P_3$ and $P_4$ in the canonical space at four different timestamps. The feature plane in \ding{193} exhibits better smoothness compared to \ding{192}, so we use \ding{193} to construct our backbone.}
  \label{fig:method_01}
\end{figure}

\begin{figure*}[ht]
  \centering
  \includegraphics[width=1\linewidth]{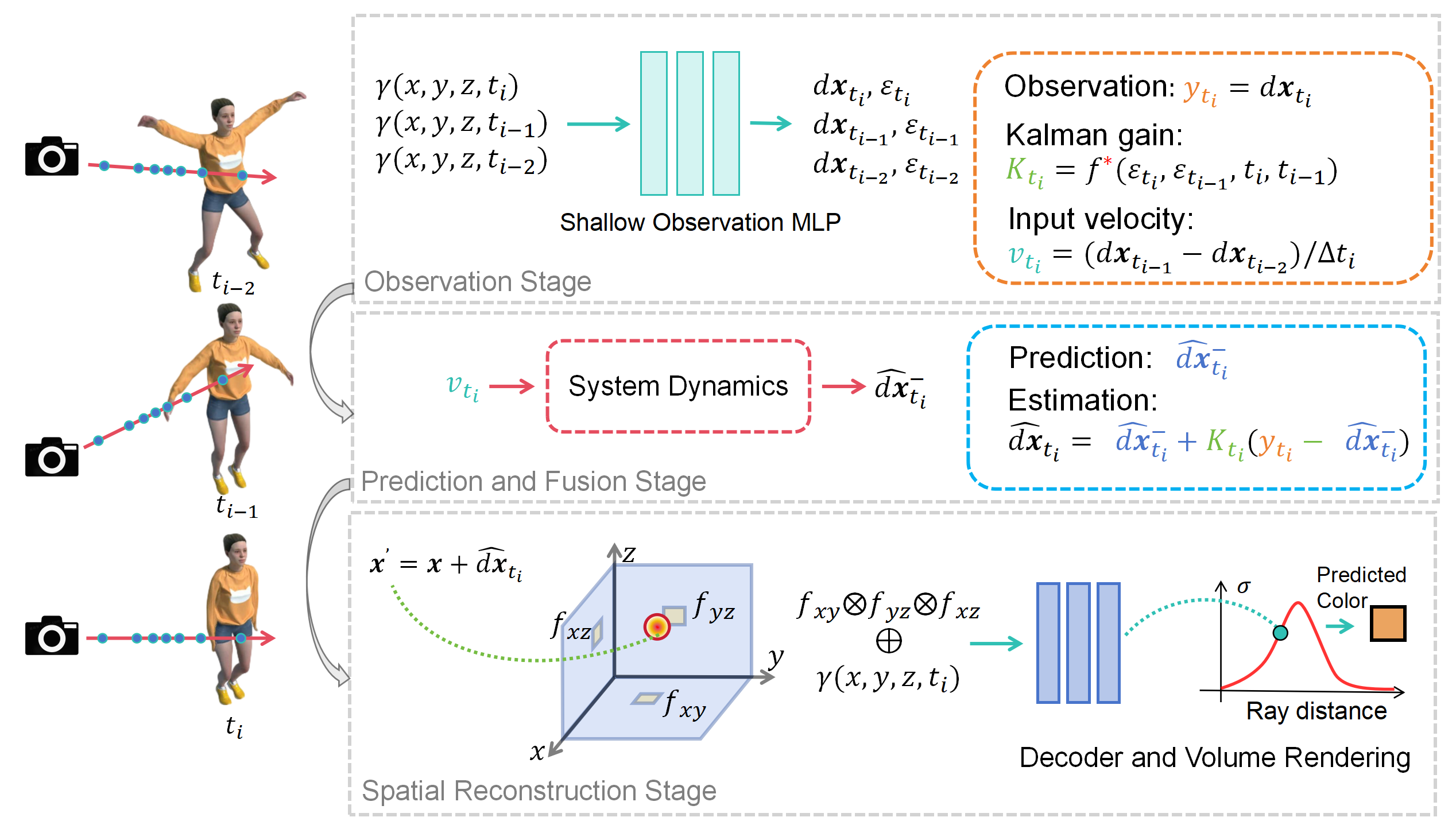}
      \caption{The overall pipeline of our KFD-NeRF. Our designed deformation field calculates final deformations based on two sources of knowledge. At the observation stage, system observations are directly output by a shallow observation MLP. At the prediction and fusion stage, we calculate predictions based on system dynamics, and further fuse observations and predictions to obtain final deformation estimations. At the spatial reconstruction stage, tri-plane encoded canonical points are concatenated with raw positions and timestamps, which are further decoded to obtain predicted colors for loss calculation. (\textcolor{red}{*}$f$ represents a linear layer, used to obtain the Kalman gain from noise-related terms $\varepsilon_{t_i}$ and $\varepsilon_{t_{i-1}}$.)}
  \label{fig:method_04}
\end{figure*}

\section{Method}
\label{sec:Method}

\Cref{fig:method_04} illustrates the three stages of the complete pipeline of KFD-NeRF. In this section, we will first analyze the advantages of using deformation fields as the motion representation over feature interpolation. We will then introduce KFD-NeRF based on the Kalman filter to achieve accurate deformation estimations. Finally, we will discuss spatial reconstruction details, the training strategy, and the incorporation of regularization.

\subsection{Motion Representation: Deformation Fields \vs Feature Interpolation}

4D NeRFs generally employ two approaches to temporal modeling: motion deformation fields or time-conditioned feature interpolation. A deformation field {$D(\bm{x},t)\rightarrow\Delta \bm{x}$} calculates the deformation $\Delta \bm{x}$ at each timestamp $t$, which can be used to warp a 3D point at $t$ to its corresponding position in the canonical space. In contrast, time-conditioned feature interpolation {$F(\bm{x},t)\rightarrow \bm{f}$} directly learns a feature vector $\bm{f}$ from a given 3D point at $t$, which is sent to a decoder for RGB and density calculation. 

In \cref{fig:method_01}, we illustrate the difference between these two methods with a motion example from time $t=0$ to time $t=1$. Given a 3D point $P$ in the real world space, we can identify its corresponding four points in the canonical space at four different timestamps, named $P_1$, $P_2$, $P_3$, and $P_4$. \ding{192} shows the feature to be learned at $P$ which changes over time. Notice that when the radiance undergoes abrupt changes, the feature space exhibits many high-frequency signals, which are hard to fit. \ding{193} shows the backward deformation from the real world space to the canonical space to be learned at $P$ which changes over time. These signals are smoother and easier to fit by leveraging the motion's smoothness prior. MLP representations have inductive smoothness bias and are well-suited for learning such deformations. Hence, we employ MLP-based deformation fields to represent motion in this work.

\subsection{Kalman Filter Guided Motion Prediction}

Next, we derive the system equations for the dynamic radiance field following \cref{eq:state equation,eq:output equation}. As we are trying to model non-rigid motions, each 3D point in the dynamic radiance field could be independently seen as a dynamic system, with its velocity $\bm{v}_{t_{i}}$ at each frame $t_{i}$ as a single input $u_t$. Also note that this dynamic system is a single-state system since we only focus on estimating deformation at each frame $t_i$, denoted as $d\bm{x}_{t_{i}}$. 

In the next step, we determine the coefficients matrix $A$, $B$, and $C$ in \cref{eq:state equation,eq:output equation}, each degenerating to a value under a single-input and single-state system. As we directly observe $d\bm{x}_{t_{i}}$, the output matrix $C$ becomes $1$. Since the motion trajectory is unknown, we use a locally-linear motion model to describe this dynamic system: $d\bm{x}_{t_{i}} = d\bm{x}_{t_{i-1}} + \Delta{t_{i}} \cdot \bm{v}_{t_{i}}$ ($A=1$ and $B=\Delta{t_{i}}$). Further considering noise, our state and output equations become
\begin{equation}
\left\{
\begin{aligned}
&d\bm{x}_{t_{i}} = d\bm{x}_{t_{i-1}} + \Delta{t_{i}} \cdot \bm{v}_{t_{i}} + n_{t_{i}},\\
&y_{t_{i}} = d\bm{x}_{t_{i}} + m_{t_{i}}\,.
\end{aligned}
\right.
\label{eq:our state and output equation}
\end{equation}

In \cref{fig:method_04}, our designed deformation field consists of two parts: an MLP-based observer and a system-dynamics-based predictor. As for the observer, by querying 3D points coordinates $\bm{x}$ and timestamps $t_{i}$, a two-layer shallow MLP is used to output the mean observation $y_{t_{i}}$ (assuming zero-mean Gaussian measurement noise $m_{t_{i}}$) and noise-related terms $\varepsilon_{t_i}$. For the predictor, we follow the locally-linear motion model and first calculate the input $\bm{v}_{t_{i}}$. Because the deformation of the current frame $t_{i}$ is being estimated, we only use the information from frame $t_{i-1}$ and frame $t_{i-2}$ to approximate the velocity
\begin{equation}
  \bm{v}_{t_i} = (d\bm{x}_{t_{i-1}}-d\bm{x}_{t_{i-2}})/\Delta{t_{i}}\,.
  \label{eq:velocity}
\end{equation}

Based on the estimated velocity, we compute the predicted deformation state $\hat{d\bm{x}}_{t_i}^-$. The final estimation $\hat{d\bm{x}}_{t_{i}}$ is the sum of $\hat{d\bm{x}}_{t_i}^-$ and $y_{t_{i}}$ weighted by a Kalman gain $K_{t_{i}}$. The Kalman gain in \cref{eq:Kaiman gain} is calculated by current measurement noise and past process noise. In our implementation, we obtain $K_{t_{i}}$ utilizing a linear layer that takes noise-related terms $\varepsilon_{t_i}$ and $\varepsilon_{t_{i-1}}$ and timestamps $t_i$ and $t_{i-1}$ as inputs, incorporating historical information. \Cref{alg:def} summarizes this deformation estimation. 

Volume is warped to the canonical space based on the estimated deformation, which is further decoded to obtain density $\sigma$ and color. We compute the loss by comparing the rendered values with the ground truth and use it to update the network, following \cref{eq:loss function}. In addition, we note that the estimated deformation serves as a good supervision for learning the observation MLP, in analogy with observation updating process in Kalman filter. Therefore, we try to minimize the error between the current observation and the estimated deformation with

\begin{equation}
  \mathcal{L}_{kf}=\frac{1}{\mathcal{N}}\sum_{\bm{x}\in\mathcal{N}}\ \left\|y_{t_{i}}-\hat{d\bm{x}}_{t_{i}}\right\|_2^2\,.
  \label{eq:kalman loss}
\end{equation}

\begin{algorithm}[t]
\caption{Deformation Estimations}
\begin{minipage}[t]{0.55\columnwidth}
\begin{algorithmic}
\STATE {\textbf{Prediction}:}

$d\bm{x}_{t_{i-1}}, \varepsilon_{t_{i-1}} = observer(\bm{x}, t_{i-1})$

$d\bm{x}_{t_{i-2}}, \varepsilon_{t_{i-2}} = observer(\bm{x}, t_{i-2})$

$\bm{v}_{t_i} = (d\bm{x}_{t_{i-1}}-d\bm{x}_{t_{i-2}})/\Delta{t_{i}}$

$\hat{d\bm{x}}_{t_{i}}^- = {d\bm{x}}_{t_{i-1}} + \Delta{t_{i}} \cdot \bm{v}_{t_i} = 2d\bm{x}_{t_{i-1}} - d\bm{x}_{t_{i-2}}$
\end{algorithmic}
\end{minipage}
\begin{minipage}[t]{0.45\columnwidth}
\begin{algorithmic}
\STATE {\textbf{Update}:}

$y_{t_{i}}, \varepsilon_{t_{i}} = observer(\bm{x}, t_{i})$

$K_{t_{i}} = f(\varepsilon_{t_{i}}, \varepsilon_{t_{i-1}}, {t_i}, {t_{i-1}})$

$\hat{d\bm{x}}_{t_{i}} = \hat{d\bm{x}}_{t_{i}}^- + K_{t_{i}}(y_{t_{i}}-\hat{d\bm{x}}_{t_{i}}^-)$
\end{algorithmic}
\end{minipage}
\label{alg:def}
\end{algorithm}

The weighted $y_{t_{i}}$ and $\hat{d\bm{x}}_{t_i}^-$ stand for two sources of knowledge from the whole system. $y_{t_{i}}$ represents the state directly observed by the observation MLP, which lacks the information from past frames and thus has a low confidence in the early stages of training. In contrast, $\hat{d\bm{x}}_{t_i}^-$ represents the current frame information predicted from history, which provides a good prior in the early stages of training. In the later stages of training, however, the confidence of the predictor drops significantly due to the inherent errors of the modeled motion dynamics, which is gradually compensated by the well learned observation MLP. Our Kalman Filter guided model automatically strikes the balance of both leveraging $\hat{d\bm{x}}_{t_i}^-$ in the early stages to accelerate convergence and avoid local minimum and also using $y_{t_{i}}$ in the later stages to learn fine details. 


\subsection{Spatial Representation}

MLP-based spatial representations suffer from slow convergence and require a deeper deformation network (\eg, eight layers MLP in~\cite{pumarola2021d}) to model 4D scenes. Otherwise, the deformation field can get stuck in a local optimum before learning the radiance field in the canonical space. Voxel grids based spatial representations converge very quickly (\eg, TiNeuVox~\cite{fang2022fast}) but demand high spatial resolution to store fine details, which requires significant memory footprint. Under memory constraints, achieving high resolution in the canonical space can be challenging, resulting in losing scene details and deteriorating the warping quality for each frame. We employ a tri-plane representation, which uses three sets of mutually orthogonal 2D planes to represent 3D space. This low-rank model allows for rapid convergence while significantly reducing memory consumption, making it possible to achieve fast and high-quality canonical space reconstruction.

What's more, there are some inevitable errors due to coordinate shifts when $\bm{x'}$ are encoded by finite resolution tri-plane. To enhance the raw coordinate information, we follow~\cite{fang2022fast} by concatenating encoded tri-plane features with raw coordinate inputs.

\subsection{Training Strategy and Regularization}

Our model takes into account temporal information so the learning of the current frame partially relies on the results of previous frames. To ensure that previous frames can offer ample priors, we employ a training strategy of gradually releasing the training images in chronological order.

We also notice that the lack of constraints or priors in the canonical space can affect the performance of the observation branch, which could be regularized by pre-setting the shape in the canonical space. D-NeRF~\cite{pumarola2021d} sets frame $t_0$ to be the canonical space and force the deformation output of frame $t_0$ to be masked to 0. Such a hard mask, however, does not allow learning of the deformation at frame $t_0$ and may cause the canonical space to be too complex for motions warping. 

We instead design a soft regularization term to normalize the difference between the canonical space and the radiance field in the real world at frame $t_0$ to improve the observation branch. The canonical observation loss of a points batch $\mathcal{N}$ is
\begin{equation}
  \mathcal{L}_{co}=\frac{1}{\mathcal{N}}\sum_{\bm{x}\in\mathcal{N}}\bm{1}(t)d\bm{x}  \,,
  \label{eq:zero canonical space loss}
\end{equation}
where $\bm{1}(t)=1$ when $t=0$ and $\bm{1}(t)=0$ when $t\ne0$.

We use a proposal sampling strategy from Mip-NeRF 360~\cite{barron2022mip} by distilling the density field for occupancy estimation. This online distillation necessitates a loss function $\mathcal{L}_{prop}$ to ensure consistency between the proposal network and our learned model. Please refer to Section $3$ in~\cite{barron2022mip} for detailed definition.

Total variation loss $\mathcal{L}_{tv}$ is a common regularizer in inverse problems, which encourages sparse edges in space. We apply this loss to each of our tri-plane to get $\mathcal{L}_{tv}(\mathbf{x'})$, where $\mathbf{x'}$ are warped 3D points in the canonical space. The total variation loss is 

\begin{small} 
\begin{equation}
\mathcal{L}_{tv}(\mathbf{x})=\frac{1}{|C|} \sum_{c, i, j}(\left\|\mathbf{x}_c^{i, j}-\mathbf{x}_c^{i-1, j}\right\|_2^2+\left\|\mathbf{x}_c^{i, j}-\mathbf{x}_c^{i, j-1}\right\|_2^2) \,,
  \label{eq:total variation loss}
\end{equation}
\end{small}

\noindent where $c$ is a certain plane from the tri-plane collection $C$ and $i$, $j$ are indices on the plane’s resolution.

The final loss function becomes

\begin{equation}
\mathcal{L}= \mathcal{L}_{image}  + 
             \mathcal{L}_{kf} + 
             \mathcal{L}_{co} + 
             \mathcal{L}_{prop} + \lambda_{tv}\mathcal{L}_{tv} \,,
  \label{eq:zero canonical space loss}
\end{equation}

\noindent and we experimentally choose $\lambda_{tv}$ to be $1\times10^{-4}$.

\section{Experimental Results}
\label{sec:Experiment}

\subsection{Dataset}

For synthetic data, we use the Dynamic NeRF Synthetic Dataset provided by D-NeRF~\cite{pumarola2021d}, whose training and testing splits have already been well-organized. For each scene in the synthetic dataset, a photo from an arbitrary view with corresponding camera pose is provided at each timestamp. For real data, we use the Nvidia Real Dynamic Scenes Dataset~\cite{yoon2020novel} which consists of 8 dynamic scenes recorded by $12$ synchronized cameras. We use $11$ camera videos for training and the remaining one for testing.

\subsection{Baselines and Metrics}
\label{subsec:Baselines and Metrics}

\begin{figure}[]
  \centering
  \includegraphics[width=0.9\linewidth]{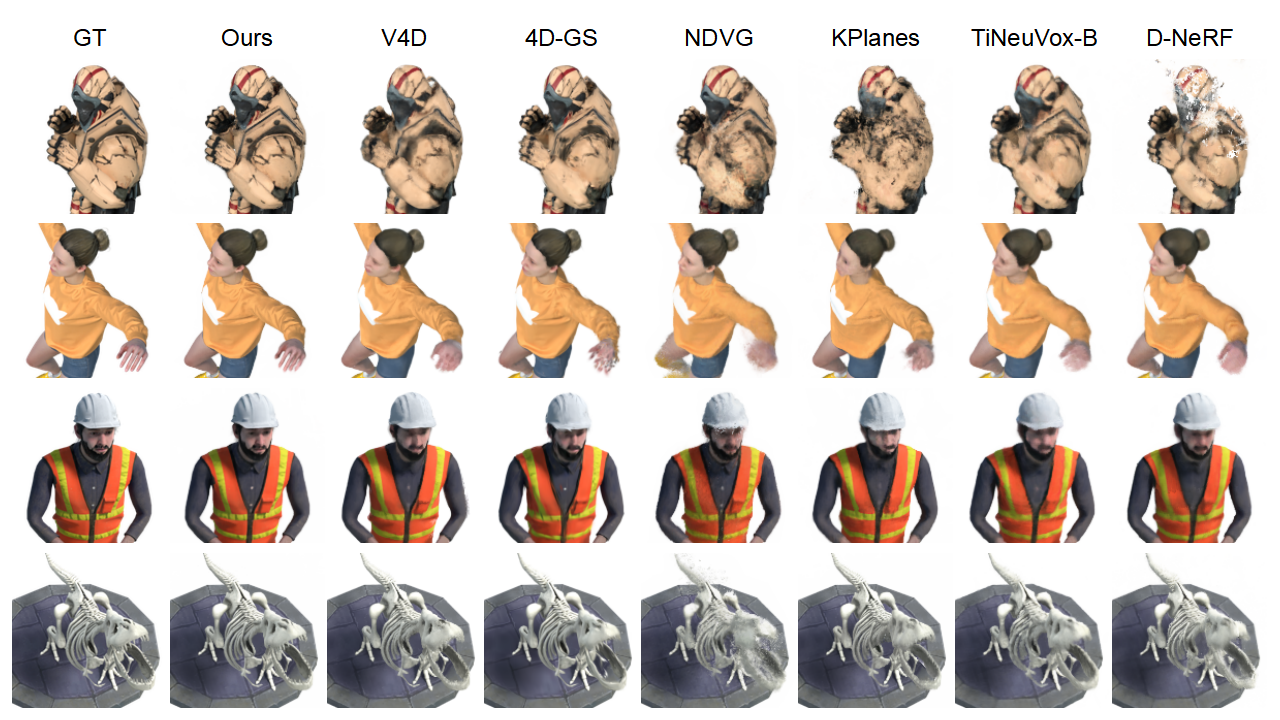}
      \caption{Qualitative Comparison of our KFD-NeRF against other dynamic NeRF methods on synthetic data. Zoom in for better details.}
  \label{fig:results_01}
\end{figure}

\begin{figure}[]
  \centering
  \includegraphics[width=1.\linewidth]{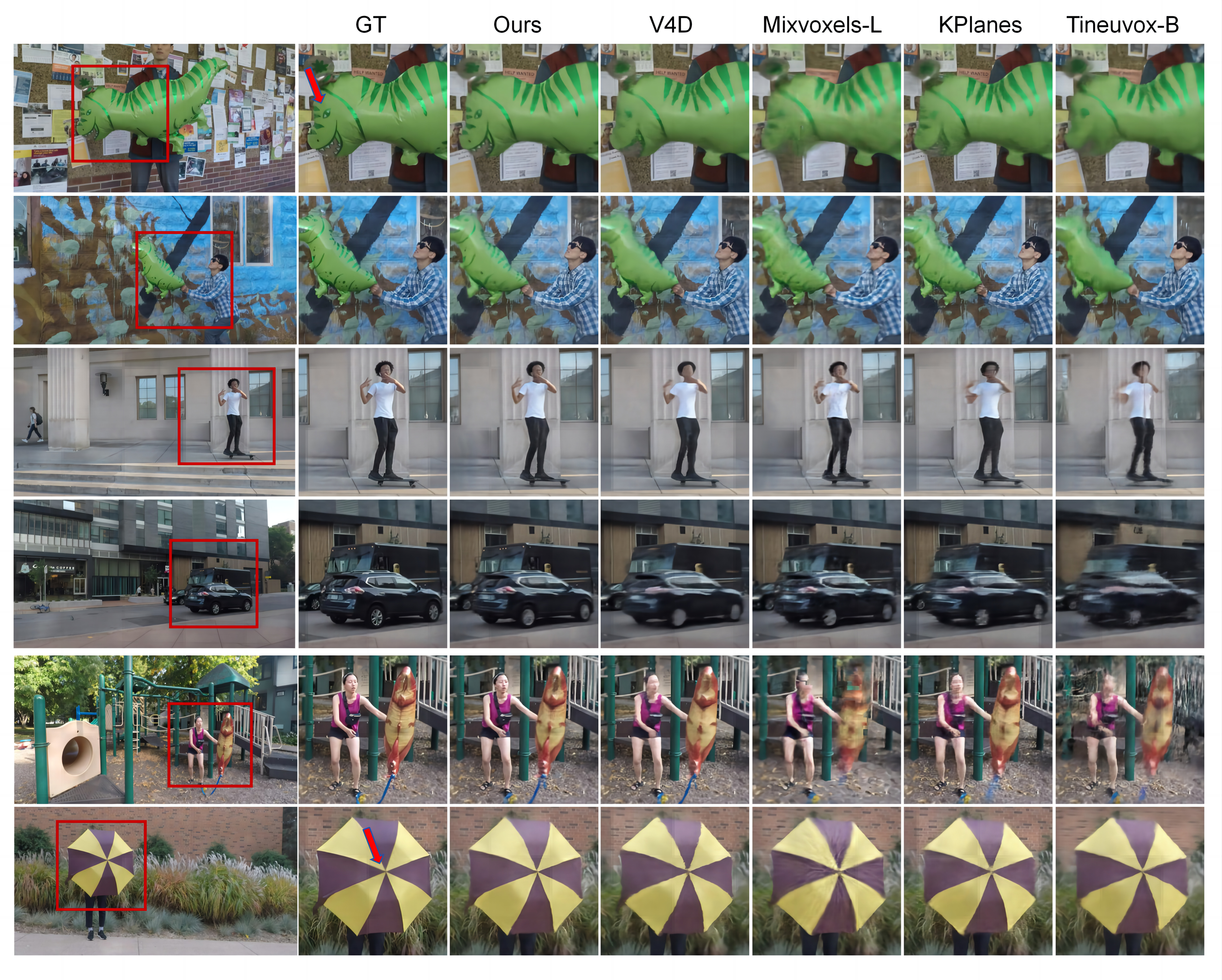}
      \caption{Qualitative Comparison of our KFD-NeRF against other dynamic NeRF methods on real data. Zoom in for better details.}
  \label{fig:results_02}
\end{figure}

\begin{table}[t]
\begingroup
\small
\begin{center}
\begin{minipage}[t]{0.49\textwidth}
\resizebox{\textwidth}{!}{
\begin{tabular}{lcccc}
\midrule
&\multicolumn{4}{c}{Synthetic Data}\\
\multicolumn{1}{l}{\textbf{Model}}       & \multicolumn{1}{c}{PSNR(dB)↑} & \multicolumn{1}{c}{SSIM↑} & \multicolumn{1}{c}{LPIPS↓} & \multicolumn{1}{c}{Average↓} \\ \cmidrule(r){1-1} \cmidrule(r){2-4} \cmidrule(r){5-5}
\multicolumn{1}{l}{D-NeRF~\cite{pumarola2021d}}      & 30.25                         & 0.954                     & 0.076                      & 0.0288                       \\
\multicolumn{1}{l}{TiNeuVox-B~\cite{fang2022fast}}    & 32.93                         & 0.973                     & 0.043                      & 0.0173                       \\
\multicolumn{1}{l}{KPlanes~\cite{fridovich2023k}}      & 30.47                         & 0.963                     & 0.072                      & 0.0261                       \\
\multicolumn{1}{l}{NDVG~\cite{guo2022neural}}        & 30.51                         & 0.964                     & 0.056                      & 0.0223                       \\
\multicolumn{1}{l}{V4D~\cite{gan2023v4d}}         & 33.54                         & 0.978                     & 0.028                      & 0.0141                       \\
\multicolumn{1}{l}{4D-GS~\cite{wu20234d}}        & 33.31                         & 0.979                     & \textbf{0.026}             & 0.0136                       \\
\multicolumn{1}{l}{KFD-NeRF}        & \textbf{35.73}                & \textbf{0.984}            & 0.045                      & \textbf{0.0131}              \\ \midrule

\end{tabular}
}
\end{minipage}
\hfill 
\begin{minipage}[t]{0.49\textwidth}
\resizebox{\textwidth}{!}{
\begin{tabular}{lcccc}
\midrule
&\multicolumn{4}{c}{Real Data}\\
\multicolumn{1}{l}{\textbf{Model}}       & \multicolumn{1}{c}{PSNR(dB)↑} & \multicolumn{1}{c}{SSIM↑} & \multicolumn{1}{c}{LPIPS↓} & \multicolumn{1}{c}{Average↓} \\ \cmidrule(r){1-1} \cmidrule(r){2-4} \cmidrule(r){5-5}
\multicolumn{1}{l}{TiNeuVox-B~\cite{fang2022fast}}  & 24.39                         & 0.734                     & 0.303                      & 0.0869                       \\
\multicolumn{1}{l}{KPlanes~\cite{fridovich2023k}}     & 28.25                         & 0.881                     & 0.137                      & 0.0433                       \\
\multicolumn{1}{l}{Mixvoxels-L~\cite{wang2023mixed}} & 26.95                         & 0.828                     & 0.222                      & 0.0619                       \\
\multicolumn{1}{l}{V4D~\cite{gan2023v4d}}         & 28.16                         & 0.884                     & 0.136                      & 0.0426                       \\
\multicolumn{1}{l}{KFD-NeRF}        & \textbf{28.75}                & \textbf{0.891}            & \textbf{0.115}             & \textbf{0.0389}              \\ \midrule
\end{tabular}
}
\end{minipage}
\end{center}
\endgroup
\caption{Quantitative comparison of our KFD-NeRF against other dynamic NeRF methods. We show results on both synthetic and real data. See \cref{subsec:Baselines and Metrics} for more details.}
\label{tab:quantitative}
\end{table}

Due to differences in the format of synthetic and real data, we carefully select cutting-edge baselines to thoroughly validate our method based on comparative experiments. For synthetic data, we test deformation based methods D-NeRF~\cite{pumarola2021d} (MLP based spatial representation), TiNeuVox-B~\cite{fang2022fast} (voxel grids based spatial representation), NDVG~\cite{guo2022neural} (voxel grids based spatial representation) and 4D-GS~\cite{wu20234d} (Gaussian points based spatial representation), and feature interpolation based methods KPlanes~\cite{fridovich2023k} and V4D~\cite{gan2023v4d}. For real data, besides TiNeuVox-B and KPlanes, we further compare multi-view videos reconstruction methods  MixVoxels~\cite{wang2023mixed}. We train all these methods on a single GeForce RTX3090. See \cref{tab:Time and Params} for detailed training time and parameters consumption of ours and other methods.

\begin{table}[t]
\begingroup
\small
\begin{center}
\resizebox{\linewidth}{!}{
\begin{tabular}{llllllllll}
\hline
\multicolumn{1}{l}{} & D-NeRF    & TiNeuVox-B & KPlanes & NDVG & V4D & 4D-GS & Mixvoxels-L & KFD-NeRF-S & KFD-NeRF-L \\ \hline
Training Time↓       &$48$hrs           &$40$mins            &$60$mins         &$35$mins      &$7$hrs     &$25$mins       &$1.5$hrs             &$30$mins            &$3$hrs            \\
Params(MB)↓              & $16$ &$50$  &$340$         &$150$      & $275$    & ~$163$      &$140$             &$175$            & $175$           \\ \hline
\end{tabular}}
\end{center}
\endgroup
\caption{Training time and parameter size for different methods.}
\label{tab:Time and Params}
\end{table}

We provide an exhaustive qualitative and quantitative comparison of our KFD-NeRF with these baseline methods. Three main metrics are reported, namely the peak signal-to-noise ratio (PSNR), the structural similarity index measure (SSIM) \cite{wang2004image}, and the learned perceptual image patch similarity (LPIPS) \cite{zhang2018unreasonable}. To provide more intuitive results, we further calculate metric ``average''~\cite{barron2021mip}, which is the geometric mean of $\rm MSE=10^{-PSNR/10}$, $\rm{\sqrt{1-SSIM}}$, and LPIPS. Please see \cref{tab:quantitative} for quantitative comparison and \cref{fig:results_01,fig:results_02} for qualitative comparison. Per-scene results can be found in the supplemental material. We strongly recommend readers to watch the supplemental videos for a more intuitive understanding of the results.

\subsection{Ablation Studies}
\label{subsec:Ablation Studies}

\begin{table}[t]
\begin{minipage}[t]{0.45\linewidth}
    \vspace{0pt}
    \centering
    \includegraphics[width=1\linewidth] {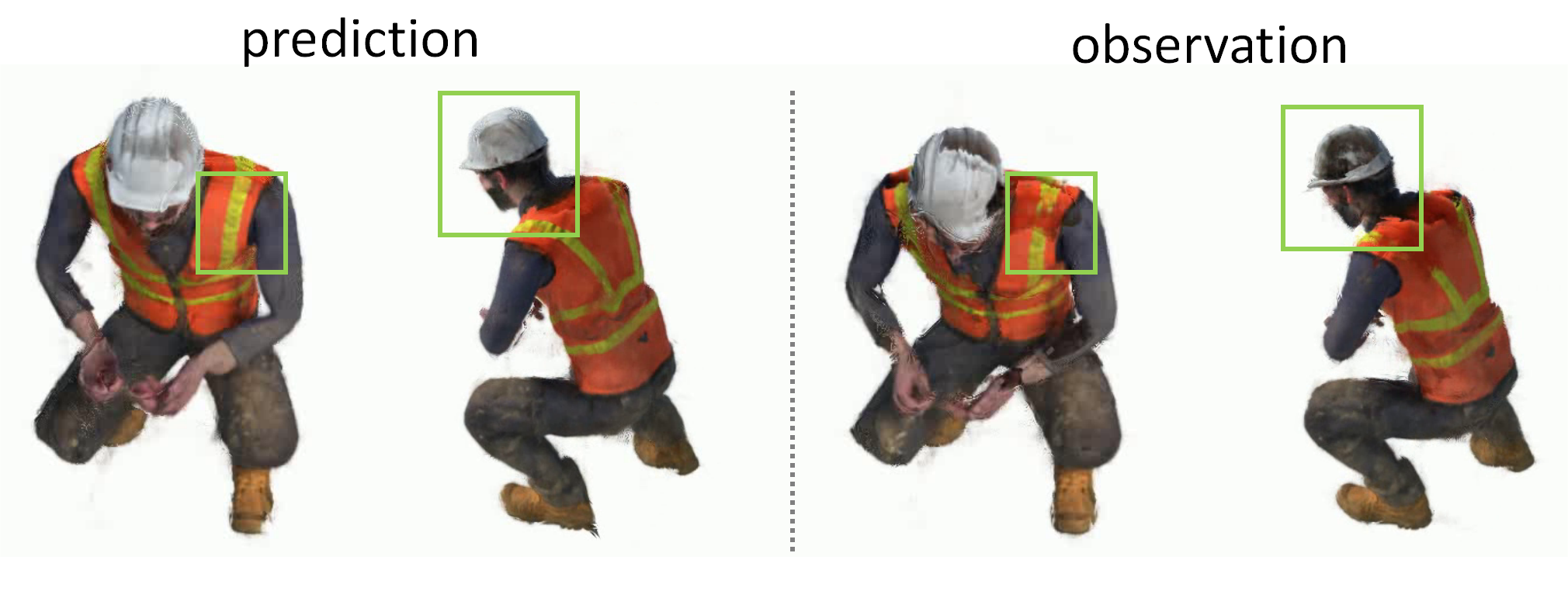}
    \captionof{figure}{Two sources of knowledge at frame $t_{i}$ from our pre-trained dynamic system with inputs up to and including frame $t_{i-1}$. Notice that during the early stages of training, our observation MLP has insufficient initialization, resulting in the loss of many details (green boxes on the right). Fortunately, our predictions provide additional priors (green boxes on the left) for the current frame based on information from previous frames, compensating for the loss in observations.}
    \label{fig:ablation_02}
\end{minipage}
\hspace{0.5cm} 
\begin{minipage}[t]{0.47\linewidth}
\vspace{0pt}
\resizebox{\linewidth}{!}{
\begin{tabular}{lcccccc}
\midrule
   & \multicolumn{1}{l}{$\mathcal{L}_{kf}$} & \multicolumn{1}{l}{$\mathcal{L}_{co}$} & PSNR(dB)↑ & SSIM↑ & LPIPS↓  \\ \cmidrule(r){1-3} \cmidrule(r){4-6}
a) &                                 &                         & 30.15     & 0.914     & 0.110           \\
b) & \Checkmark                                &                         & 30.60     & 0.920     & 0.106  \\
c) &                                 & \Checkmark                        & 31.78     & 0.931     & 0.086   \\
$d)$ & \Checkmark                                & \Checkmark                       & \textbf{32.24}     & \textbf{0.938}     & \textbf{0.080}\\ \midrule
\end{tabular}}
\captionof{table}{Ablation Studies on the $\mathcal{L}_{kf}$ and $\mathcal{L}_{co}$. We show all the ablation combinations and report the average results on synthetic and real data.}
\label{tab:ablation}
\vspace{-0.2cm}

\resizebox{\linewidth}{!}{
\begin{tabular}{lcccc}
\midrule
&\multicolumn{2}{c}{Dynamic} & \multicolumn{2}{c}{Static}\\
\multicolumn{1}{l}{\textbf{Model}}       & \multicolumn{1}{c}{PSNR(dB)↑} & \multicolumn{1}{c}{SSIM↑} & \multicolumn{1}{c}{PSNR(dB)↑} & \multicolumn{1}{c}{SSIM↑} \\ \cmidrule(r){1-1} \cmidrule(r){2-3} \cmidrule(r){4-5}
\multicolumn{1}{l}{P.N., $(\bm{x_{i}}, t_{i})$ Input}  & 25.61                         & 0.963                     & \textbf{29.58}                      & 0.812                       \\
\multicolumn{1}{l}{P.N., $(\bm{x_{i}}, t_{i}, t_{i-1}, t_{i-2})$ Input}     & 25.66                         & 0.964                     & 29.55                     & 0.812                       \\
\multicolumn{1}{l}{w/ Prediction Branch (full)}        & \textbf{26.71}                & \textbf{0.967}            & 29.57            & \textbf{0.813}              \\ \midrule

\end{tabular}
}
\captionof{table}{The ablation study on our prediction branch for real data in average, where dynamic and static areas are counted separately. P.N. is short for Pure MLP Network.}
\label{tab:Prediction Branch ablation}

\end{minipage}
\end{table}

We conduct ablation studies on both synthetic and real data to validate our various proposed system components. We compare our full model with variants related to $\mathcal{L}_{co}$, $\mathcal{L}_{kf}$ and prediction branch.

\noindent \textbf{Canonical observation loss $\mathcal{L}_{co}$}. This loss is designed to regularize a continuous and smooth volume shape in the canonical space for better observations. Specifically, we guide the shape in the canonical space to be close to the shape at frame $t_0$. In \cref{tab:ablation} line $b)$, we remove $\mathcal{L}_{co}$ and observe a decrease in model performance.

\noindent \textbf{Estimation update loss $\mathcal{L}_{kf}$}. This loss tries to minimize the difference between estimated deformation and current observation. This updating process ensures that with each iteration, the observation acquire progressively more accurate information to guide the learning process. In \cref{tab:ablation} line $c)$, we remove $\mathcal{L}_{kf}$ and witness a decrease in model performance.

\subsection{Extra Studies on Prediction Branch}

This study aims to demonstrate the effectiveness of our plug-in Kalman filter. The most crucial step in Kalman filter is to fuse the original network observations with the predictions from system dynamics. Therefore, we directly remove the ``Prediction and Fusion Stage'' in our pipeline and use pure shallow observation MLP to generate deformations. We further conduct an experiment where a pure deformable network takes the current frame and the previous two states as inputs to ablate the effectiveness of the Kalman motion modeling. In \cref{tab:Prediction Branch ablation}, we see a clear drop in performance by pure MLP network-based methods, indicating that simply inputting previous Kalman filter parameters without modeling motion prediction is insufficient.

The prediction branch we have designed can offer reliable priors in the early stages of training. We visualize these priors through an experiment to gain further insight. Specifically, we train our system with only inputs up to and including frame $t_{i-1}$ and then let our system produce rendering results at frame $t_{i}$ with two different branches: predictions and observations. This experiment effectively simulates the amount of knowledge contained in predictions and observations in the early stages of training when the system encounters new input data. In \cref{fig:ablation_02} we show two sources of knowledge from prediction branch and from direct output of observation MLP at frame $t_{i}$. We see an obvious deficiency in the observation MLP when initializing new input frames and are pleased to find that the prediction branch compensates for this loss based on information from previous frames.

\section{Discussion and Conclusion}
\label{sec:Discussion and Conclusion}

\begin{figure}[t]
  \centering
  \includegraphics[width=1\linewidth]{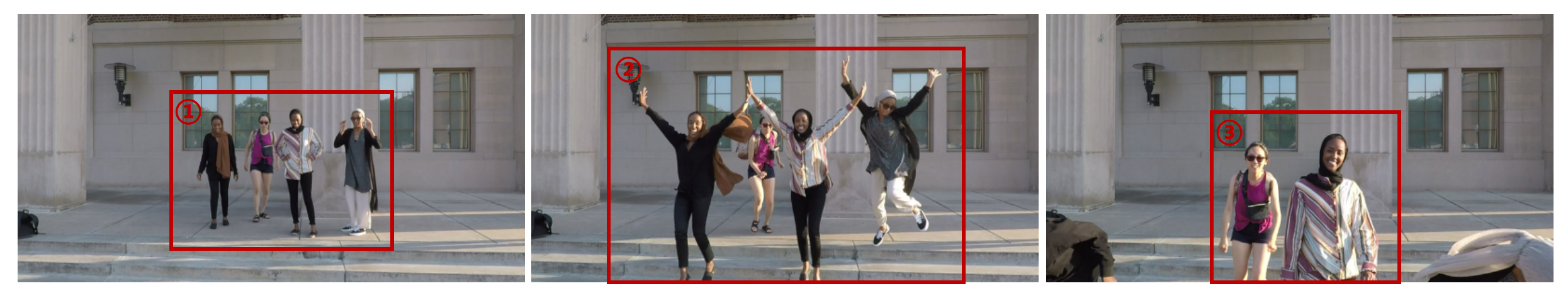}
      \caption{An example in Nvidia Dataset~\cite{yoon2020novel} to illustrate scale and topological issues in choosing canonical space. From \ding{192} to \ding{193} there is a scale change due to moving closer to the camera. From \ding{194} to \ding{193} there is a topological change caused by target disappearance. According to this, choosing canonical space \ding{192} will suffer from low resolution and choosing canonical space \ding{194} will suffer from incomplete topology while \ding{193} is a better canonical space to choose.}
  \label{fig:ablation_01}
\end{figure}

\noindent\textbf{Limitations}. Our method relies on a well-reconstructed radiance field in the canonical space, which in our pipeline is guided by $\mathcal{L}_{co}$. This design, however, will partially fail, if the chosen canonical space exhibits significant scale changes or even topological changes related to other frames. In ~\cref{fig:ablation_01} we use an example to illustrate the influences caused by choosing canonical space. This phenomenon, however, cannot be mitigated by precisely setting the canonical space as we lack a priori access to the input data. We note that some works~\cite{barron2021mip, park2021hypernerf} focus on addressing scale or topological issues in radiance fields reconstruction.
Nevertheless, these issues are not the main focus of our paper and will be explored in future works to further refine our model.

\noindent\textbf{Conclusion}. In this work, we present KFD-NeRF, a Kalman filter guided NeRF, for 4D dynamic view synthesis. We model the dynamic radiance field as a dynamic system in control theory and use Kalman filter to estimate the deformation states based on both observations and predictions. We further enhance our observation by encoding canonical space with an efficient tri-plane and by regularizing the shape in the canonical space. Through our temporal training strategy and newly derived pipeline, KFD-NeRF achieves state-of-the-art view synthesis performance among a variety of dynamic NeRF methods.
We hope the dynamic system modeling of 4D radiance fields will encourage researchers to explore motion contextual information. KFD-NeRF hopefully inspires the use of existing sequential methods mainly in control theory and visual state estimation to further improve 4D view synthesis and deformation estimations tasks.

\section*{Acknowledgements}

This research was supported in part by JSPS KAKENHI Grant Numbers 24K22318, 22H00529, 20H05951, JST-Mirai Program JPMJMI23G1.

%
%
\bibliographystyle{splncs04}
\bibliography{main}
\end{document}


\title{KFD-NeRF: Rethinking Dynamic NeRF with Kalman Filter\\
Supplemental Material}


\author{
Yifan Zhan\inst{1}
\and
Zhuoxiao Li\inst{1}
\and
Muyao Niu\inst{1}
\and
Zhihang Zhong\inst{3}
\and
Shohei Nobuhara\inst{2}
\and
Ko Nishino\inst{2}
\and
Yinqiang Zheng\thanks{\footnotesize denotes corresponding author.}\inst{1}
}

\authorrunning{Y.Zhan, Z.Li et al.}

\institute{
The University of Tokyo
\and
Kyoto University
\and
Shanghai Artificial Intelligence Laboratory
}

\begin{figure}[h]
\maketitle
\centering
\thispagestyle{empty}
\centering
\includegraphics[width=\linewidth]{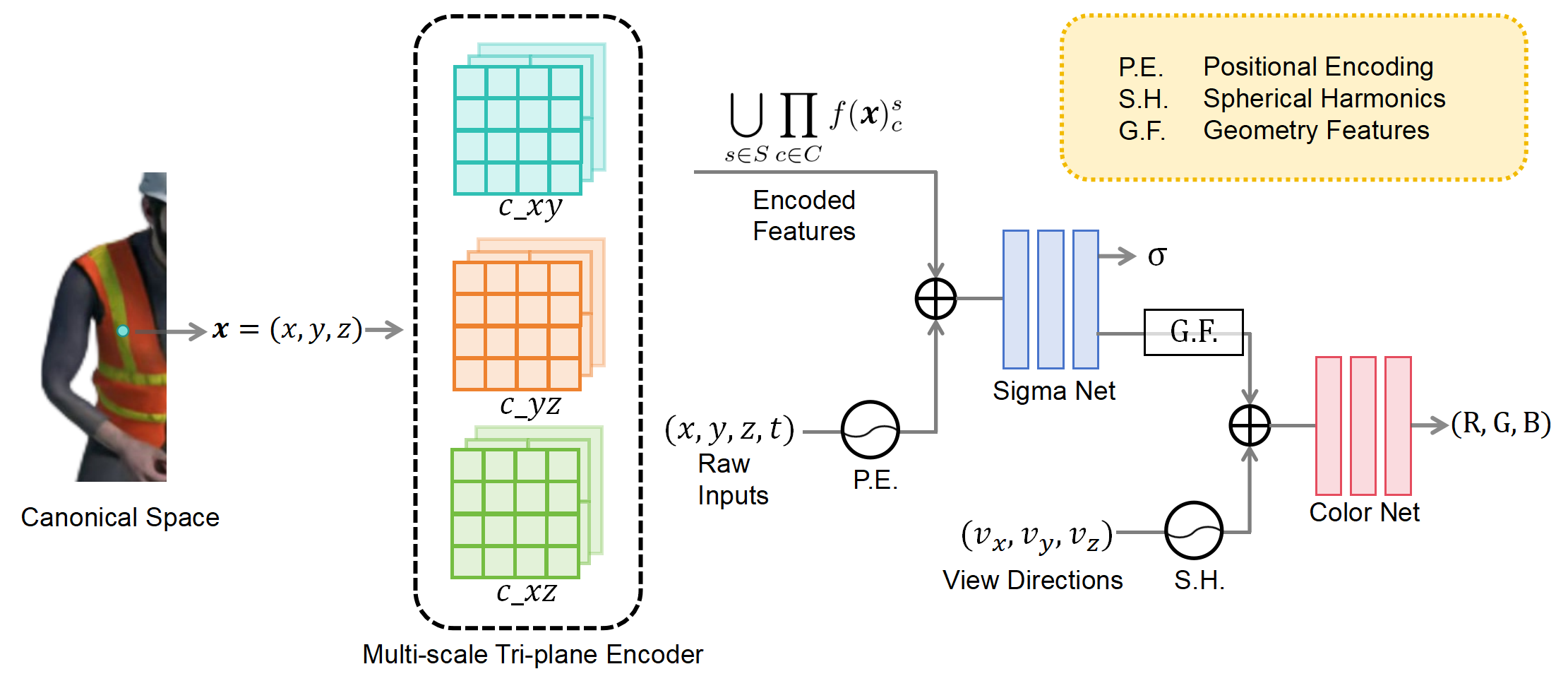}
\captionof{figure}{Details of our spatial neural architectures.}
\label{fig:details of spatial}
\end{figure}

\section{Per-scene Results}
\label{sec:Per-scene Results}

\begin{table*}[ht]
\begingroup
\small
\begin{center}
\resizebox{\linewidth}{!}{
\begin{tabular}{lcccccccccccccccc}
\hline
         & \multicolumn{4}{c}{Bouncing Balls}                                & \multicolumn{4}{c}{Hell Warrior}                                  & \multicolumn{4}{c}{Hook}                                          & \multicolumn{4}{c}{Jumping Jacks}                                 \\
\textbf{Method}   & PSNR↑          & SSIM↑          & LPIPS↓         & Average↓           & PSNR↑          & SSIM↑          & LPIPS↓         & Average↓           & PSNR↑          & SSIM↑          & LPIPS↓         & Average↓           & PSNR↑          & SSIM↑          & LPIPS↓         & Average↓           \\ \cmidrule(r){1-1} \cmidrule(r){2-5} \cmidrule(r){6-9} \cmidrule(r){10-13} \cmidrule(r){14-17}
D-NeRF~\cite{pumarola2021d}   & 38.05          & 0.982          & 0.107          & 0.018          & 24.94          & 0.948          & 0.071          & 0.038          & 29.43          & 0.962          & 0.122          & 0.032          & 31.96          & 0.972          & 0.045          & 0.019          \\
TiNeuVox-B~\cite{fang2022fast} & 40.53          & 0.990          & 0.036          & 0.007          & 28.21          & 0.965          & 0.072          & 0.027          & 32.31          & 0.975          & 0.044          & 0.016          & 34.70          & 0.983          & 0.033          & 0.012          \\
KPlanes~\cite{fridovich2023k}   & 41.12          & 0.991          & 0.032          & 0.006          & 25.65          & 0.952          & 0.079          & 0.036          & 28.55          & 0.957          & 0.082          & 0.029          & 32.60          & 0.977          & 0.054          & 0.017          \\
NDVG~\cite{guo2022neural}     & 34.59          & 0.969          & 0.113          & 0.019          & 25.58          & 0.949          & 0.075          & 0.037          & 29.74          & 0.966          & 0.040          & 0.021          & 29.55          & 0.960          & 0.081          & 0.027          \\
V4D~\cite{gan2023v4d}      & 42.00          & \textbf{0.992} & \textbf{0.029} & \textbf{0.005} & 27.06          & 0.960          & 0.054          & 0.028          & 30.95          & 0.972          & 0.037          & 0.018          & 35.29          & 0.986          & \textbf{0.022} & \textbf{0.010} \\
4D-GS~\cite{wu20234d}     & 39.18          & 0.990          & 0.033          & 0.007          & 27.53          & 0.968          & \textbf{0.042} & 0.024          & 32.13          & 0.978          & \textbf{0.022} & 0.013          & 34.71          & 0.986          & \textbf{0.022} & \textbf{0.010} \\
KFD-NeRF     & \textbf{42.16} & 0.991          & 0.038          & 0.006          & \textbf{30.58} & \textbf{0.978} & 0.051          & \textbf{0.019} & \textbf{35.65} & \textbf{0.990} & 0.034          & \textbf{0.010} & \textbf{35.64} & \textbf{0.988} & 0.041          & 0.011 \\ \hline
         & \multicolumn{4}{c}{Lego}                                          & \multicolumn{4}{c}{Mutant}                                        & \multicolumn{4}{c}{Stand Up}                                      & \multicolumn{4}{c}{T-Rex}                                         \\
\textbf{Method}   & PSNR↑          & SSIM↑          & LPIPS↓         & Average↓           & PSNR↑          & SSIM↑          & LPIPS↓         & Average↓           & PSNR↑          & SSIM↑          & LPIPS↓         & Average↓           & PSNR↑          & SSIM↑          & LPIPS↓         & Average↓           \\ \cmidrule(r){1-1} \cmidrule(r){2-5} \cmidrule(r){6-9} \cmidrule(r){10-13} \cmidrule(r){14-17}
D-NeRF~\cite{pumarola2021d}   & 21.70          & 0.842          & 0.169          & 0.077          & 31.19          & 0.973          & 0.029          & 0.016          & 33.41          & 0.980          & 0.023          & 0.012          & 31.30          & 0.972          & 0.043          & 0.018          \\
TiNeuVox-B~\cite{fang2022fast} & 25.17          & 0.924          & 0.075          & 0.040          & 33.76          & 0.979          & 0.031          & 0.013          & 36.03          & 0.985          & 0.021          & 0.009          & 32.74          & 0.979          & 0.033          & 0.014          \\
KPlanes~\cite{fridovich2023k}   & 25.52          & \textbf{0.948} & 0.059          & 0.034          & 24.71          & 0.917          & 0.178          & 0.057          & 33.89          & 0.983          & 0.051          & 0.014          & 31.71          & 0.981          & 0.038          & 0.016          \\
NDVG~\cite{guo2022neural}     & 25.21          & 0.933          & 0.052          & 0.034          & 35.44          & 0.988          & 0.015          & 0.008          & 34.01          & 0.982          & 0.023          & 0.011          & 30.00          & 0.967          & 0.048          & 0.021          \\
V4D~\cite{gan2023v4d}      & \textbf{25.63} & \textbf{0.948} & \textbf{0.038} & \textbf{0.029} & 36.14          & 0.989          & 0.014          & 0.007          & 37.06          & 0.990          & 0.012          & \textbf{0.006} & 34.21          & 0.987          & \textbf{0.018} & \textbf{0.010} \\
4D-GS~\cite{wu20234d}     & 25.37          & 0.940          & 0.044          & 0.031          & 37.80          & 0.993          & \textbf{0.009} & \textbf{0.005} & 36.82          & 0.990          & \textbf{0.011} & 0.007          & 32.97          & 0.984          & 0.022          & 0.012          \\
KFD-NeRF     & 25.54          & 0.948          & 0.070          & 0.035          & \textbf{39.23} & \textbf{0.995} & 0.039          & 0.007          & \textbf{39.62} & \textbf{0.994} & 0.030          & 0.007          & \textbf{37.40} & \textbf{0.992} & 0.054          & \textbf{0.010} \\ \hline
\end{tabular}
}
\end{center}
\endgroup
\caption{Per-scene quantitative comparisons on synthetic dynamic scenes.}
\label{tab:per-scene quantitative synthetic}
\end{table*}

\begin{table*}[ht]
\begingroup
\small
\begin{center}
\resizebox{\linewidth}{!}{
\begin{tabular}{lcccccccccccccccc}
\hline
                & \multicolumn{4}{c}{Balloon1}                                      & \multicolumn{4}{c}{Balloon2}                                      & \multicolumn{4}{c}{dynamicFace}                                   & \multicolumn{4}{c}{Jumping}                                       \\
\textbf{Method} & PSNR↑          & SSIM↑          & LPIPS↓         & Average↓           & PSNR↑          & SSIM↑          & LPIPS↓         & Average↓           & PSNR↑          & SSIM↑          & LPIPS↓         & Average↓           & PSNR↑          & SSIM↑          & LPIPS↓         & Average↓           \\ \cmidrule(r){1-1} \cmidrule(r){2-5} \cmidrule(r){6-9} \cmidrule(r){10-13} \cmidrule(r){14-17}
TiNeuVox-B~\cite{fang2022fast}        & 25.21          & 0.773          & 0.249          & 0.071          & 26.35          & 0.814          & 0.208          & 0.059          & 22.30          & 0.887          & 0.167          & 0.069          & 25.41          & 0.779          & 0.329          & 0.077          \\
KPlanes~\cite{fridovich2023k}          & 28.20          & 0.887          & 0.100          & 0.037          & 26.85 & 0.863          & 0.157          & 0.044 & 25.44          & 0.923          & 0.112          & 0.045          & 27.09          & 0.857          & 0.206          & 0.058          \\
Mixvoxels-L~\cite{wang2023mixed}       & 26.24          & 0.808          & 0.235          & 0.063          & 26.78          & 0.811          & 0.235          & 0.060          & 20.03          & 0.792          & 0.308          & 0.112          & 26.91          & 0.855          & 0.230          & 0.059          \\
V4D~\cite{gan2023v4d}             & 27.11          & 0.888          & 0.101          & 0.040          & 24.55          & 0.847          & 0.148          & 0.059          & \textbf{27.20} & \textbf{0.951} & \textbf{0.083} & \textbf{0.033} & \textbf{27.78} & \textbf{0.883} & \textbf{0.175} & \textbf{0.049} \\
KFD-NeRF            & \textbf{28.83} & \textbf{0.906} & \textbf{0.076} & \textbf{0.031} & \textbf{27.30}          & \textbf{0.888} & \textbf{0.089} & \textbf{0.038} & 26.45          & 0.937          & 0.092          & 0.037          & 26.93          & 0.845          & 0.215          & 0.061          \\ \hline
                & \multicolumn{4}{c}{Playground}                                    & \multicolumn{4}{c}{Skating}                                       & \multicolumn{4}{c}{Truck}                                         & \multicolumn{4}{c}{Umbrella}                                      \\
\textbf{Method} & PSNR↑          & SSIM↑          & LPIPS↓         & Average↓           & PSNR↑          & SSIM↑          & LPIPS↓         & Average↓           & PSNR↑          & SSIM↑          & LPIPS↓         & Average↓           & PSNR↑          & SSIM↑          & LPIPS↓         & Average↓           \\ \cmidrule(r){1-1} \cmidrule(r){2-5} \cmidrule(r){6-9} \cmidrule(r){10-13} \cmidrule(r){14-17}
TiNeuVox-B~\cite{fang2022fast}        & 16.60          & 0.376          & 0.461          & 0.200          & 27.93          & 0.840          & 0.276          & 0.056          & 25.78          & 0.765          & 0.356          & 0.077          & 25.56          & 0.636          & 0.381          & 0.086          \\
KPlanes~\cite{fridovich2023k}          & 24.59          & 0.836          & 0.152          & 0.060          & 34.06          & 0.956          & 0.092          & 0.020          & 32.93          & 0.925          & 0.118          & 0.025          & 26.84          & 0.801          & 0.159          & 0.057          \\
Mixvoxels-L~\cite{wang2023mixed}       & 23.19          & 0.748          & 0.253          & 0.085          & 33.14          & 0.945          & 0.148          & 0.026          & 32.60          & 0.920          & 0.133          & 0.027          & 26.70          & 0.748          & 0.236          & 0.063          \\
V4D~\cite{gan2023v4d}             & \textbf{25.69} & \textbf{0.875} & \textbf{0.100} & \textbf{0.046} & 33.79          & 0.956          & 0.115          & 0.022          & 32.95          & 0.930          & 0.126          & 0.026          & 26.24          & 0.742          & 0.237          & 0.066          \\
KFD-NeRF            & 24.59          & 0.846          & 0.129          & 0.056          & \textbf{34.95} & \textbf{0.964} & \textbf{0.077} & \textbf{0.017} & \textbf{33.44} & \textbf{0.930} & \textbf{0.100} & \textbf{0.023} & \textbf{27.49} & \textbf{0.815} & \textbf{0.146} & \textbf{0.048} \\ \hline
\end{tabular}
}
\end{center}
\endgroup
\caption{Per-scene quantitative comparisons on real dynamic scenes.}
\label{tab:per-scene quantitative real}
\end{table*}

\begin{table}[!h]
\begingroup
\small
\begin{center}
\resizebox{0.8\linewidth}{!}{
\begin{tabular}{lccccc}
\hline
\textbf{Model}   & PSNR(dB)↑      & SSIM↑          & LPIPS↓         & Training Time↓  & Params(MB)↓  \\ \cmidrule(r){1-1} \cmidrule(r){2-4} \cmidrule(r){5-5} \cmidrule(r){6-6}
MLP Based        & 32.13          & 0.971          & 0.052          & 43hrs           & \textbf{0.5} \\
Voxel Grid Based & 36.21          & 0.984          & \textbf{0.029}          & \textbf{60mins} & 2336         \\
Tri-plane Based  & \textbf{38.92} & \textbf{0.990} & 0.037 & 150mins         & 31.5         \\ \hline
\end{tabular}
}
\end{center}
\endgroup

\caption{\textbf{Ablation Study on Spatial Representations}. We compare three different spatial models on rendering quality, training time and spatial-only size. The tri-plane model we choose significantly reduces the model size compared to the voxel grid based model and achieves the highest rendering quality in a relatively short training time.}
\label{tab:ablation on space}
\end{table}

\begin{figure*}[]
  \centering
  \includegraphics[width=1\linewidth]{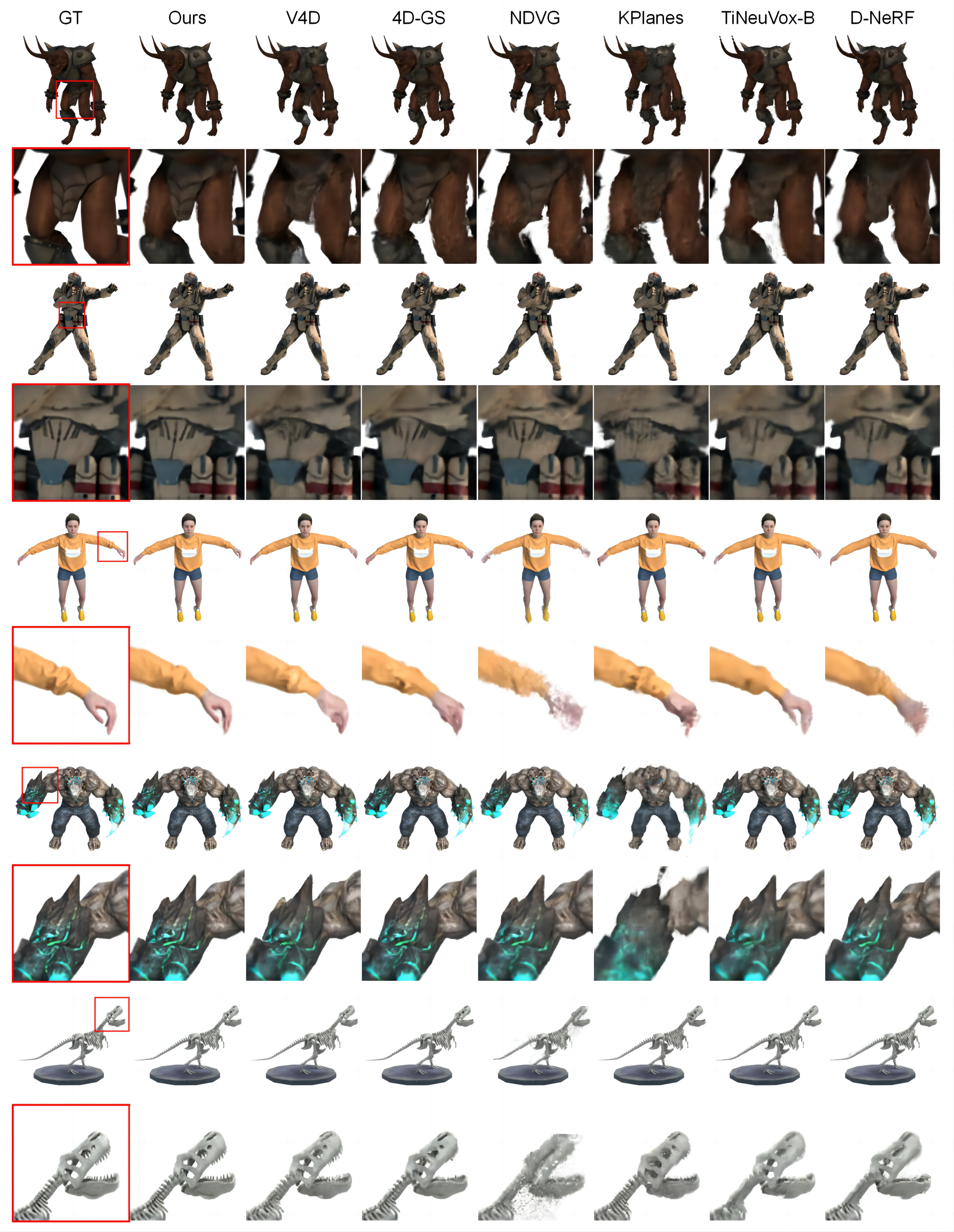}
      \caption{More qualitative results on synthetic data. Zoom in for better details.}
  \label{fig:asupplemental material_234}
\end{figure*}

\begin{figure*}[]
  \centering
  \includegraphics[width=0.9\linewidth]{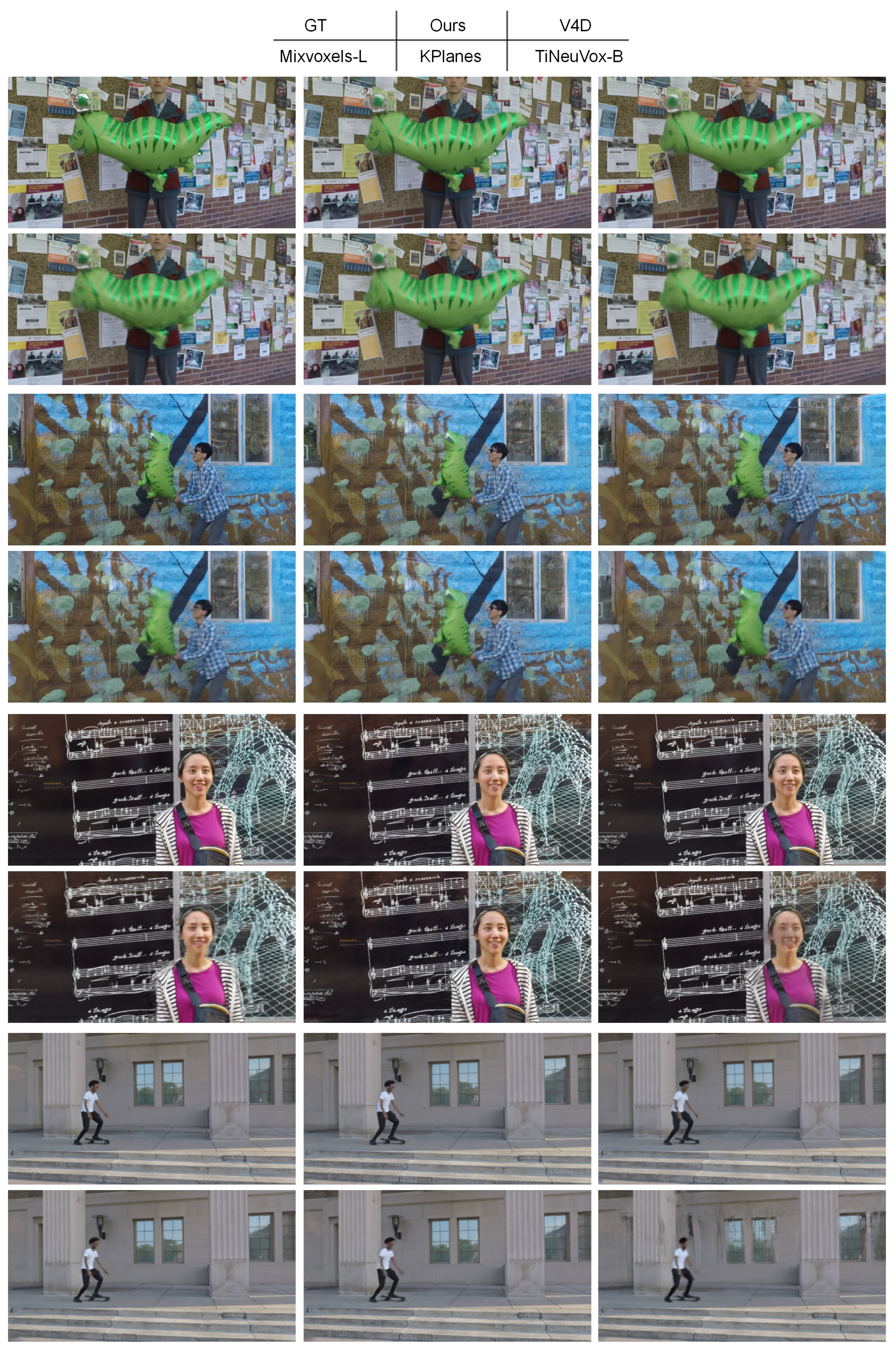}
      \caption{More qualitative results on real data. Zoom in for better details.}
  \label{fig:asupplemental material_567}
\end{figure*}

We exhibit our per-scene results for synthetic data in \cref{tab:per-scene quantitative synthetic} and for real data in \cref{tab:per-scene quantitative real}. In \cref{fig:asupplemental material_234} and \cref{fig:asupplemental material_567} we additionally provide more visualization results on synthetic and real data.

\section{Spatial Neural Architectures}
\label{sec:Neural Architectures}

We elaborate in detail on our spatial neural architectures in \cref{fig:details of spatial}. Once obtaining the warped 3D points in the canonical space, we first use a multi-scale tri-plane to encode the spatial information. For each plane $c\in{C}$ and each scale $s\in{S}$, features are multiplied across planes and concatenated across scales to obtain tri-plane encoded features. However, coordinate shifts happen due to the limited resolution of the tri-plane grids and errors introduced by linear interpolation. We follow~\cite{fang2022fast} by concatenating encoded tri-plane features with positional encoded~\cite{rahaman2019spectral} raw inputs. Our Sigma Net (single-hidden-layer MLP) outputs volume density $\sigma$ and 15-dimensional geometry features. The geometry features will be further concatenated with spherical hamonics encoded view directions for color calculation using Color Net (two-hidden-layer MLP).

\section{Model Hyperparameters}
\label{sec:Model Hyperparameters}

The frequency number of positional encoding is set to $5$ for both spatial and temporal inputs. Our shallow observation MLP consists of two hidden layers, each has a channel dimension of $128$. As for the tri-plane grids, we use multi-scale planes with $4$ different resolutions at $64^2$, $128^2$, $256^2$ and $512^2$. As can be seen in \cref{fig:details of spatial}, the per-plane and per-scale features are multiplied across planes and concatenated across scales. We set each of these per-plane and per-scale features' dimension to be $32$ so the final encoded feature has dimension $128$. Sigma Net has single hidden layer with channel dimension $64$ and Color Net has two hidden layers with channel dimension $64$.

For optimization, an Adam optimizer~\cite{kingma2014adam} is used and we set ray batch to be $4096$ in each iteration. We train our KFD-NeRF with learning rate set to $1\times10^{-3}$.

\section{Ablation Study on Spatial Representations}
\label{sec:More Ablation Studies}

In Section $4.3$, we analyze the characteristics of different spatial representations and their reconstruction capabilities in the canonical space. Based on our full model, we further conduct ablation studies by only changing spatial representations to show their impacts on reconstruction results.

Specifically, we compare three different spatial models, namely, pure MLP (8 hidden layers with dimension $256$), multi-scale voxel grid (resolutions at $64^3$, $128^3$ and $256^3$) and multi-scale tri-plane (resolutions at $64^2$, $128^2$ and $256^2$). In \cref{tab:ablation on space}, We compare these models based on three dimensions: model size, convergence time, and rendering quality. 

Our full model only uses a shallow MLP with two hidden layers to calculate deformations, so the pure spatial MLP would suffer from convergence difficulties. The voxel grid based model converges faster while the model size is far from acceptable. Therefore we  choose the tri-plane based model as a spatial model that converges relatively fast, has a moderate size, and provides high rendering quality.

%
%
\bibliographystyle{splncs04}
\bibliography{main}